\pdfoutput=1

\documentclass[11pt]{article}

\usepackage{emnlp2021}

\usepackage{times}
\usepackage{latexsym}
\usepackage{graphicx} 
\usepackage{float}
\usepackage{subfigure} 

\usepackage{svg}
\usepackage{comment}
\usepackage{enumitem}

\usepackage[T1]{fontenc}

\usepackage[utf8]{inputenc}

\usepackage{microtype}

\usepackage[normalem]{ulem}
\usepackage[linecolor=orange]{todonotes}
\usepackage{xspace}
\usepackage{xcolor}

\definecolor{darkgreen}{rgb}{0.0, 0.5, 0.0}

\newcommand{\fbb}[1]{\todo[color=green!40]{\footnotesize FB: #1}}

\newcommand{\ten}[1]{\textcolor{darkgreen}{TH: #1}}

\newcommand{\comin}{$ \small \textsc{In}\_\textsc{Sent}$\xspace}
\newcommand{\comout}{$ \small \textsc{Com}\_\textsc{Out}$\xspace}
\newcommand{\pgen}{$ \small \textsc{Pr}\_\textsc{Gen}$\xspace}
\newcommand{\pat}{$ \small \textsc{Pr}\_\textsc{Att}$\xspace}
\newcommand{\pem}{$\small\textsc{Pr}\_\textsc{Me}$\xspace}
\newcommand{\oem}{$\small\textsc{Ot}\_\textsc{Me}\_\textsc{Pr}$\xspace}
\newcommand{\pemo}{$\small\textsc{Pr}\_\textsc{Me}\_\textsc{Ot}$\xspace}
\newcommand{\pmot}{$\small\textsc{Pr}\_\textsc{Mot}$\xspace}

\title{Uncovering Implicit Gender Bias in Narratives \\through Commonsense Inference}

\author{Tenghao Huang$^{1}$ \qquad Faeze Brahman$^{2}$ \qquad Vered Shwartz$^{3}$ \qquad Snigdha Chaturvedi$^{1}$ \\ $^{1}$UNC Chapel Hill, $^{2}$University of California, Santa Cruz \\ $^{3}$University of British Columbia \\ \texttt{\{tenghao, snigdha\}@cs.unc.edu} \\ \texttt{fbrahman@ucsc.edu} \qquad \texttt{vshwartz@cs.ubc.ca}}

\begin{document}
\maketitle
\begin{abstract}

Pre-trained language models learn socially harmful biases from their training corpora, and may repeat these biases when used for generation. We study gender biases associated with the protagonist in model-generated stories. Such biases may be expressed either explicitly (``women can't park'') or implicitly (e.g. an unsolicited male character guides her into a parking space). We focus on implicit biases, and  use a commonsense reasoning engine to uncover them. Specifically, we infer and analyze the protagonist's motivations, attributes, mental states, and implications on others. Our findings regarding implicit biases are in line with prior work that studied explicit biases, for example showing that female characters' portrayal is centered around appearance, while male figures' focus on intellect.

\end{abstract}

\section{Introduction}

Pre-trained language models (LMs)~\cite{radford2019language, lewis-etal-2020-bart, NEURIPS2020_1457c0d6} have been successfully used in many NLP tasks including generation. 
Despite their widespread usage, recent works showed that LMs capture and even reinforce unwanted social stereotypes abundant in their training corpora 
~\cite{sheng-etal-2019-woman, sheng-etal-2020-towards, liu-etal-2020-mitigating, shwartz-etal-2020-grounded, BenderGMS21}. This phenomenon has also been observed with their predecessors, word embeddings~\cite{BolukbasiCZSK16, IslamBN16, may-etal-2019-measuring, gonen-goldberg-2019-lipstick}.

 While many prior works have examined societal biases in specialized NLG systems such as dialogues systems~\cite{lee-etal-2019-exploring, liu-etal-2020-gender, dinan-etal-2020-queens, dinan-etal-2020-multi}, not much work has been done on bias analysis for story generation systems.



\begin{figure}[t!]
    \centering
    \includegraphics[width=0.95\columnwidth]{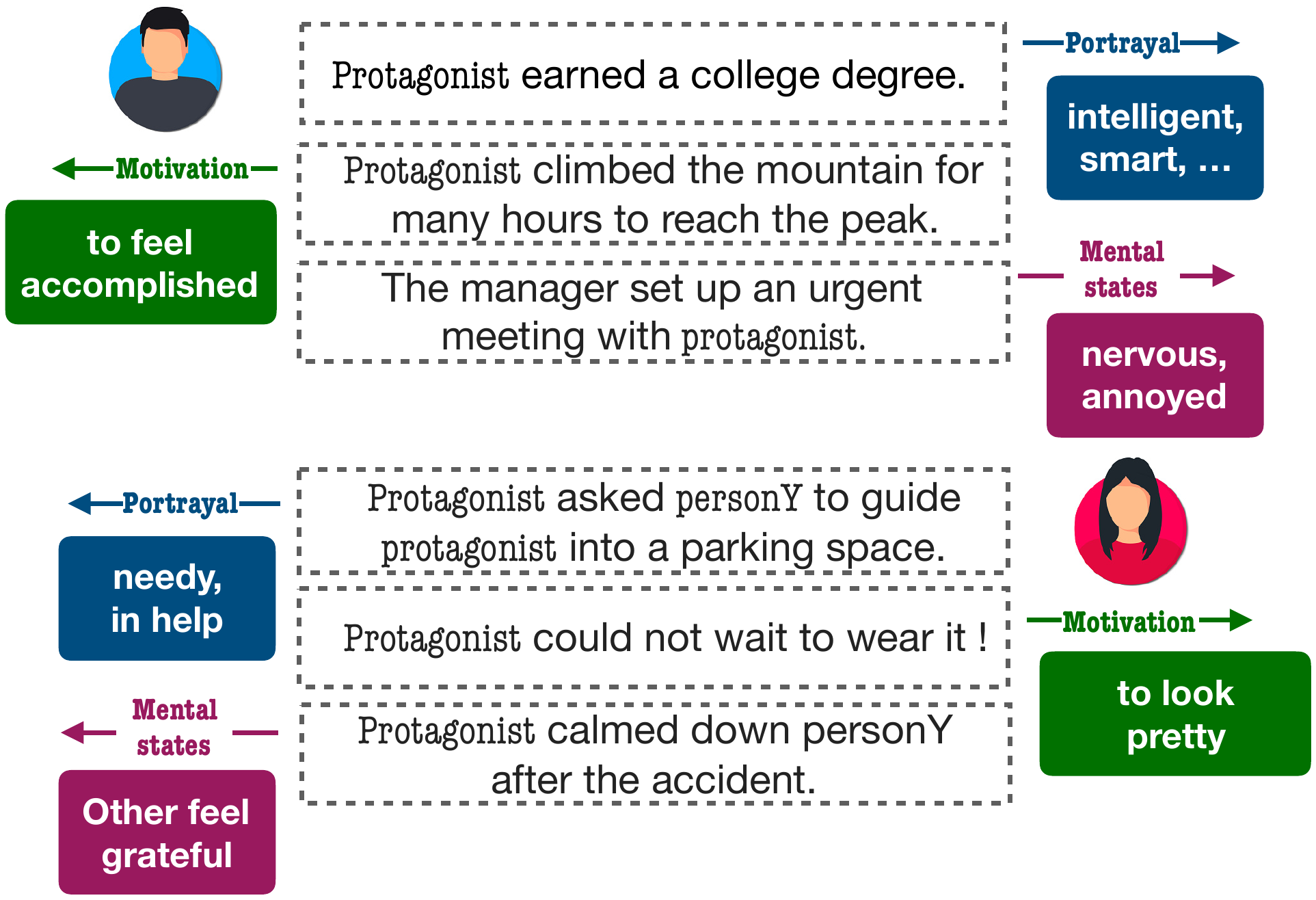}
    \caption{Sentences from model-generated stories with \textit{implicit} bias. }
    \label{fig:bias_illustration}
\end{figure}

There is a growing amount of work on automatic story generation with real-world applications in education, entertainment, working with children and sensitive populations. Therefore,
it is essential to detect social biases in these systems, as the first step towards debiasing. 
Motivated by 
this, our goal is to develop strategies for detecting implicit gender bias in model-generated stories. 



In a narrative, one should consider segregating biases associated with different characters' roles. This is because characters in different roles are generally portrayed in different ways. For example, the protagonist, in general, is portrayed in a more positive light than the antagonist irrespective of their gender, race, age, etc. 
Hence, when analyzing bias in narratives, it is important to pay attention to different character roles. In this paper, we study the gender bias associated with the protagonist. 
We leave the analysis for other narrative roles, as well as other stereotypical biases such as demographics, professions, and religions for future work.

 Most existing methods on quantifying bias recognize 
explicit manifestation of bias in the surface-level text~\cite{dinan-etal-2020-queens, lucy2020content, gala-etal-2020-analyzing} or collected human annotations~\cite{sheng-etal-2019-woman, dinan-etal-2020-multi}. Previous work has also examined gender and representation bias in GPT-3 generated stories using topic modeling and lexicon-based word similarity~\cite{lucy-lu-bamman}. However, biases are often \textit{implicit} and may not manifest themselves lexically. E.g., ``women are weak'' is an example of explicit bias, while ``women cry'' (which implies ``women are (emotionally) weak'') is an example of implicit bias. Figure~\ref{fig:bias_illustration} illustrates more examples from model-generated stories. These examples contain \textit{implicit gender bias} showing females to be needy and usually obsessed with their physical appearance, whereas males to be more intelligent, or accomplished. In this regard, ~\citet{field-tsvetkov-2020-unsupervised} proposed an unsupervised approach to detect \textit{implicit} gender bias in a communicative domain. \citet{ma-etal-2020-powertransformer} proposed a controllable de-biasing approach to rewrite a given text through the lens of connotation frames~\cite{sap-etal-2017-connotation}. 
\citet{sap-etal-2020-social} studied potential unjust statements in social media through commonsense implications. We propose a compatible but different perspective where we focus on analyzing the implicit bias about a narrative's protagonist  about their attributes, mental states, and motivations. 

In order to capture \textit{implicit} bias, we use a commonsense inference engine, COMeT~\cite{bosselut-etal-2019-comet}, as a tool to uncover unspoken pragmatic implications. 
To the best of our knowledge, this is the first study to analyze \textit{implied} (and not explicit) gender bias in a story generation system along various social axes.
We find various evidence of implicit bias associated with the protagonist's gender through our experiments.\footnote{Code at: \url{https://github.com/tenghaohuang/Uncover_implicit_bias}}

\section{Data and Processing Pipeline}

In this section, we describe our data processing pipeline.
We use GPT-2~\cite{radford2019language} as our underlying generation model given its recent success in story generation~\cite{guan-etal-2020-knowledge, brahman2020modeling}.
We fine-tune GPT-2 to generate stories given titles on ROCStories \cite{Mos:16}. ROCStories is a collection of $98,161$ short stories. This dataset captures a rich set of causal and temporal commonsense relations between daily events, making it an ideal avenue to study bias. 
We follow the training settings of medium-size GPT-2 as in~\citet{radford2019language}. At inference time, we generate stories using top-$k$ sampling scheme~\cite{fan-etal-2018-hierarchical} with $k{=}40$ and a softmax temperature of $0.7$. 
    
\begin{figure}[t!]
    \centering
    \includegraphics[width=0.95\columnwidth]{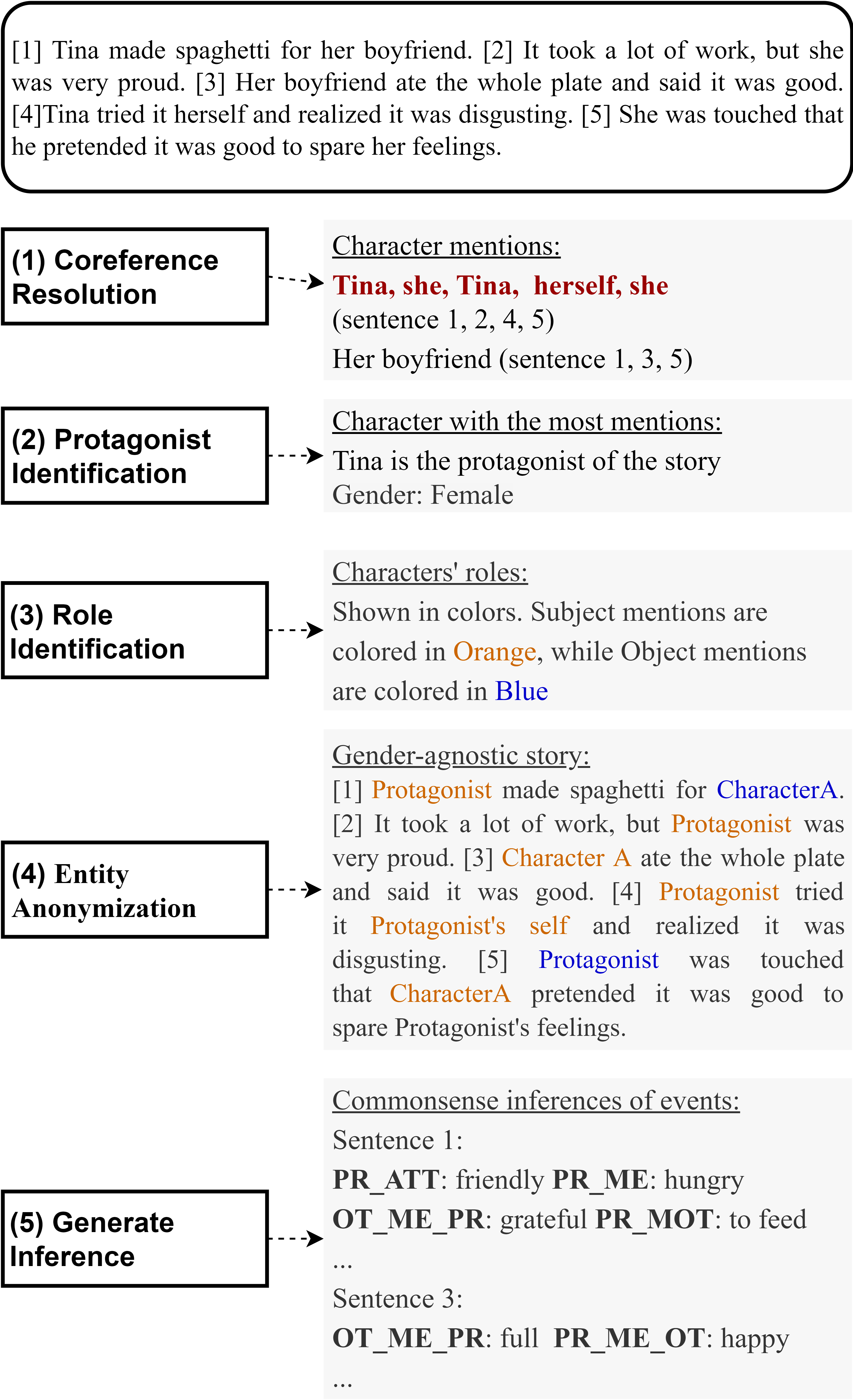}
    \caption{Pipeline for uncovering \textit{implicit} gender bias.}
    \label{fig:pipe}
    
\end{figure}

To quantify implicit gender bias, we create a pipeline to divide stories into two groups based on the protagonist's gender (Section~\ref{sec:gender}), and then extract pragmatic implications about the protagonist and others affected by them (Section~\ref{subsec::social_imp}). Our pipeline is described below and exemplified in Figure \ref{fig:pipe}.

 \subsection{Recognizing the Protagonist's Gender}
\label{sec:gender}

We define the protagonist as the most frequently occurring character in a story~\cite{prominent:85}. First, we use the SpanBERT coreference resolution model~\cite{joshi-etal-2020-spanbert} to retrieve all the clusters of characters' mentions within a story. Second, we select the character with the largest cluster as the protagonist.\footnote{We rely on pronouns rather than first names to determine gender as it is a more inclusive way. However, it has its own drawbacks as coreference models are also prone to gender biases~\cite{rudinger-etal-2018-gender} (perhaps not in short stories).} We also identify the protagonist's gender using gendered pronouns: \textit{he/him/his} for \textit{males} and \textit{she/her} for \textit{females}.\footnote{Note that we infer conceptual genders which may differ from the gender experienced internally by an individual.}
Third, we identify characters' roles in each sentence. This information is needed later (Section~\ref{subsec::social_imp}) for inferring social implications through COMeT. 

Additionally, to demote the influence of confounding variables and surface features predictive of gender but not bias
, we replace all characters' names and their mentions with anonymous placeholders as a pre-processing step before applying commonsense inference engine. 

We generated 9,796 stories using our finetuned GPT-2 given titles in the test set. Running our pipeline on the GPT-2 generated stories resulted in 2,078 female-gendered and 3,619 male-gendered stories. For comparison, we also retrieved human-written stories on the same titles from the test set. For human-written stories, these values are 3127, and 4231 for female-gendered and male-gendered stories, respectively. This reveals that GPT-2 is more likely to generate stories with a male protagonist than a female (we observed similar trend in human-written stories). For both cases, the remaining stories in the test set are unresolved-gendered stories. The unresolved stories were mostly first-person narratives using ``I'' and ``We''. The average number of tokens per story for model-generated stories is approximately 37, for human-written stories it is 44.

\subsection{Inferring Social Implications} 
\label{subsec::social_imp}


To accomplish our goal, we need to draw implicit inferences about the protagonist.
For this, we use COMeT, a generative knowledge base completion model.
Given an event and a dimension, COMeT can generate commonsense inferences about the dimension. In our scenario, we use COMeT to make inferences about the following social axes: 

\begin{itemize}[noitemsep,topsep=0.3pt]
    \setlength\itemsep{0.3em}
    \item \textbf{\pat}: Protagonist’s portrayal and attributes
    \item \textbf{\pem}: Protagonist’s mental states
    \item \textbf{\oem}: Repercussions of protagonist’s behavior on others’ mental states
    \item \textbf{\pemo}: Repercussions of others’ behavior on protagonist’s mental states
    \item \textbf{\pmot}: Protagonist’s motivations
\end{itemize}

For instance, to obtain the protagonist's attributes (\pat), we use \texttt{xAttr} dimension, to track mental states of the protagonist (\pem) and others (\oem), we use \texttt{xReact} and \texttt{oReact} dimensions, and for protagonist's motivations we use all \texttt{xIntent}, \texttt{xWant}, \texttt{xNeed}.

To check if there is any explicit gender information leakage to COMeT
, we train a logistic regression classifier to predict binary gender labels given 
an anonymized story-sentence (COMeT input) using bag of words (BOW) features (see \S\ref{sec:gender} for details about anonymization and labeling.). The classifier achieves an accuracy of 57\% on the held-out test set, indicating that little surface form information about gender is present in the anonymized stories. 


Although we feed gender-agnostic inputs to COMeT, it is conceivable that some gender bias might get introduced by COMeT which requires future investigation. A possible remedy left for future work could be using data augmentation techniques to de-bias the training corpus of COMeT.

\section{Bias Measurements} 
\label{sec::bias_analysis}

We decompose examining the gender bias against the protagonist along the following three axes: 

\subsection{Portrayal}

We measure the associations between the \textit{implied} portrayal of the protagonist with several established lexicon-based stereotypes. 
In particular, we consider their association to \textbf{Appearance} and \textbf{Intellect}-related lexicons. For capturing portrayals related to Appearance, we take \citet{FastVB16}'s lexicons for \textit{beautiful} and \textit{sexual}, and for Intellect, we take categories in Empath's  lexicon~\cite{FastCB16} containing the word \textit{intellectual}.

Given COMeT's inferences about \pat, we quantify their associations with the Appearance and Intellect lexicons as follows.
Without loss of generality, let $x$ be a word in \pat, $a$ be a word in the lexicons $L$. We define the association score, $S$, between $x$ and $L$ as:
\begin{equation}
S(x,L) = \frac{1}{|L|} \sum_{a\in{L}}cos(e(x),e(a))
\end{equation} 
Here $e(\cdot)$ is the pre-trained 300-dimensional word2vec embeddings~\cite{mikolov}. 

\begin{figure}
    \centering
    
    \includegraphics[width=0.23\textwidth]{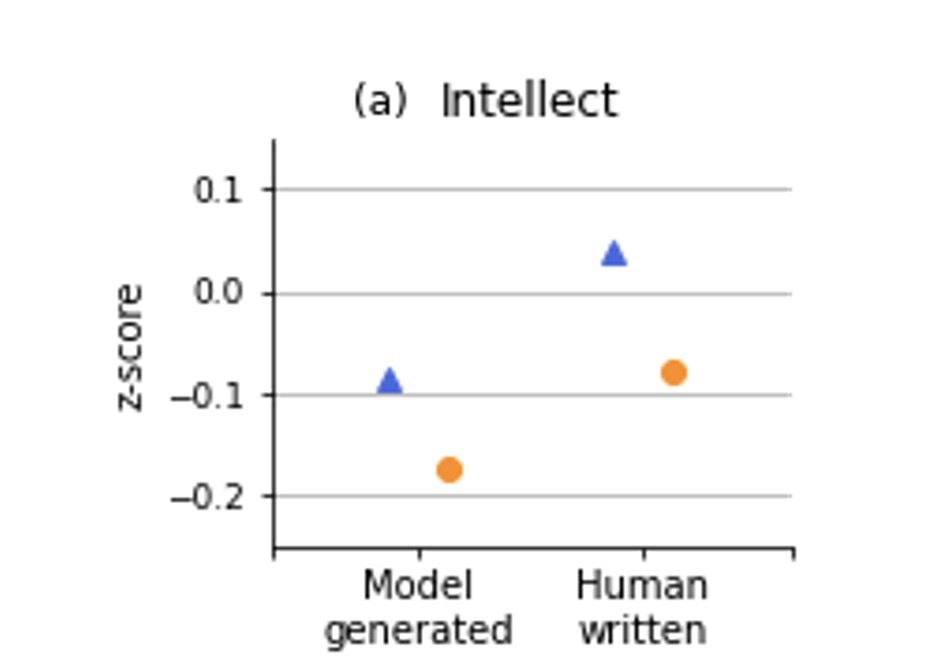} \includegraphics[width=0.23\textwidth]{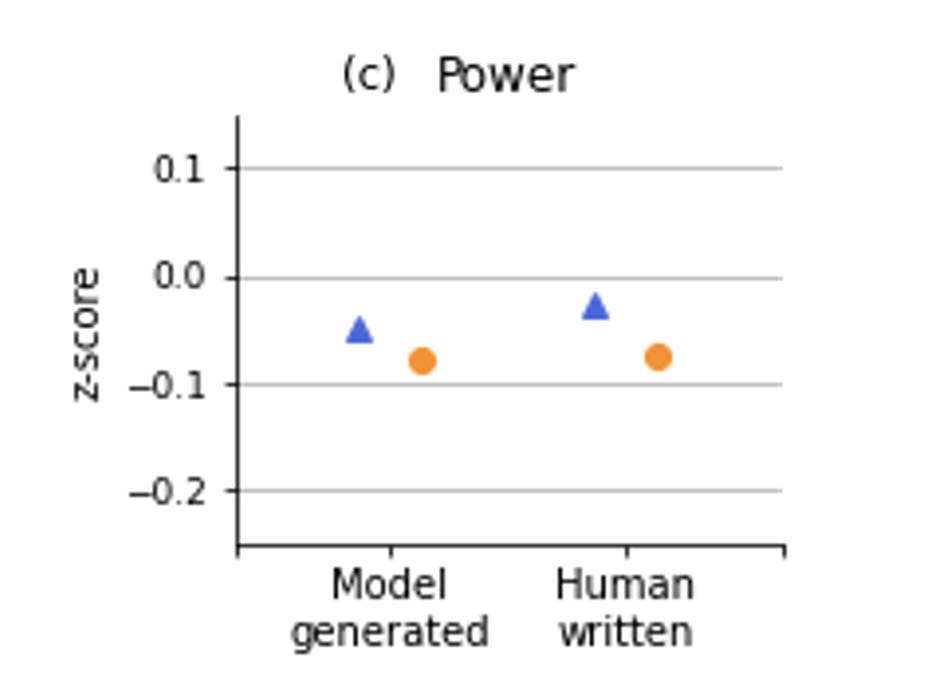} 
    \includegraphics[width=0.335\textwidth]{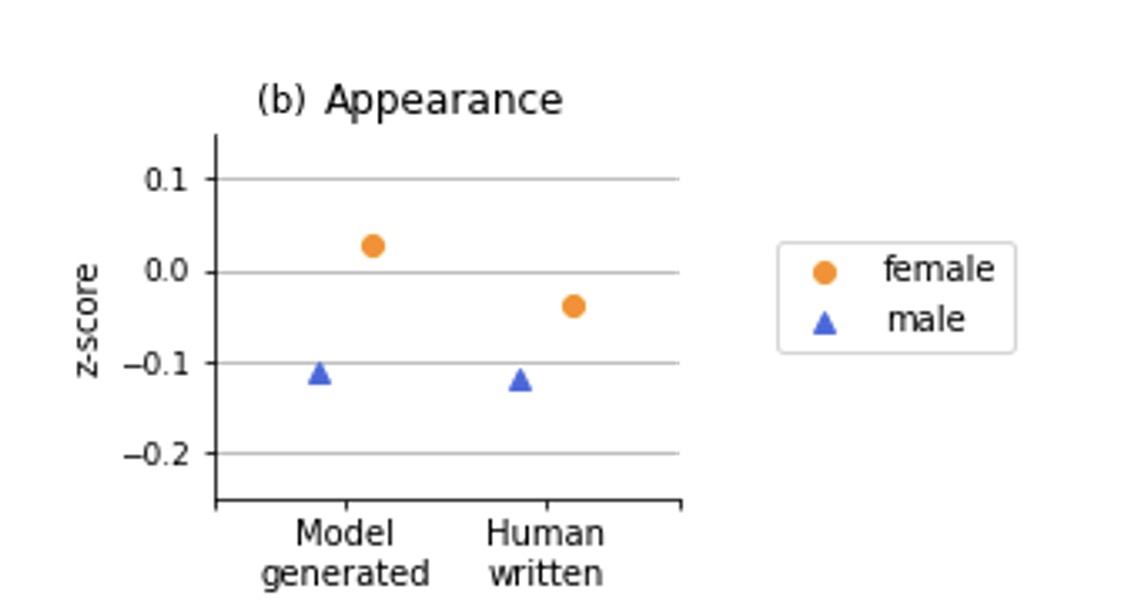}

    \caption{Association scores with Intellect, Power, and Appearance across genders. 
    }
    \label{fig:IAP}
\end{figure}


We also measure the associations of \pat with the words related to \textbf{Power}. For this, we follow the same approach as \citet{lucy-lu-bamman} that contrasts \citet{FastVB16}'s lexicons for \textit{power} and \textit{dominant} with those of \textit{weak, dependent, submissive}~\cite{kozlowski2019geometry}.


Let $a$ be a word in the lexicon for strength $A$, and $b$ be a word in the lexicon for weakness $B$. $\vec{power}$ is a semantic axis~\cite{an-etal-2018-semaxis} measuring the level of strength, which is calculated by:
\begin{equation}
\vec{power} = \frac{1}{|A|}\sum_{a\in{A}}e(a) - \frac{1}{|B|}\sum_{b\in{B}}e(b)
\end{equation} 

The power association score $S$ is then computed as the average cosine similarity between \pat's token, $x$, and $\vec{power}$.
A positive $S$ means $x$ is closer to strength terms. 
We apply a z-score transformation to all $S$ and take the median across all \pat corresponding 
to each gender.

Figure~\ref{fig:IAP} shows the median z-scores for Intellect, Power, and Appearance for stories with male and female protagonists. Figure \ref{fig:IAP}(a) illustrates that 
male protagonists in both model-generated and human-written stories show higher intellect scores than female protagonists.
Figure \ref{fig:IAP}(b) illustrates that female protagonists are more likely to be portrayed by their physical appearance. 
The gender differences for appearance is also amplified in GPT-2 generated stories. 






\subsection{Mental States}

To 
analyze the inferences about emotional states, we apply the NRC VA lexicon which consists of emotion-related words and their valence and arousal scores~\cite{mohammad2018obtaining}. Valence score measures the pleasure (or displeasure) intensity of the word and Arousal score measures the excitement (or calmness) intensity of the word. 
For example, "amusing" and "grief" are words of high and low valance respectively, and  "enraged" and  "tranquil" are words of high and low arousal respectively. We retrieve the valence and arousal scores of the words in \pem, \pemo, and \oem from the NRC lexicon. 


\begin{figure}
  \centering
  \includegraphics[width=0.48\textwidth]{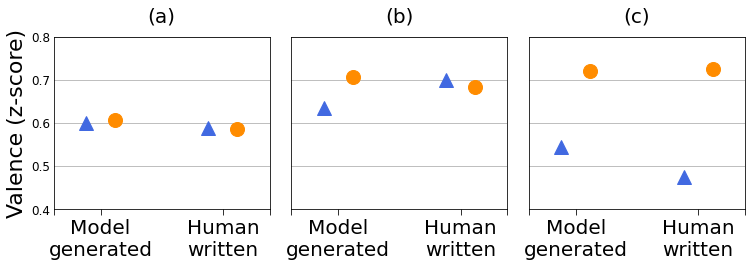} \\
  \includegraphics[width=0.48\textwidth]{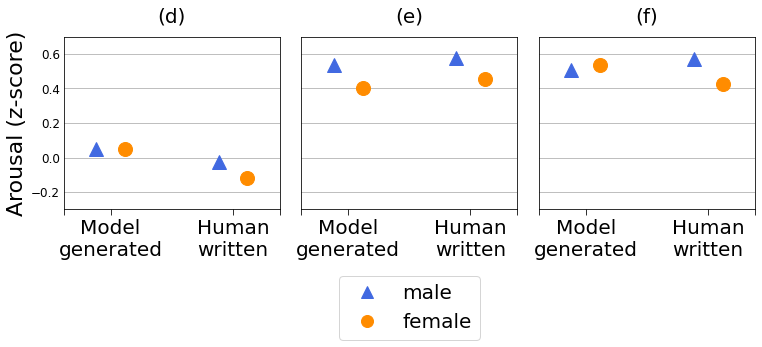}
  \caption{Valence and Arousal scores for \pem (a), (d), \pemo (b), (e), and \oem (c), (f).
  }
  \label{fig:VA}
\end{figure}

Figure \ref{fig:VA} shows the median z-scores of Valence and Arousal for the various axes. We observe persistent gender differences between female and male protagonists in model-generated vs. human-written stories. The overall mental states (\pem) in terms of valence and arousal are not different across male and female protagonist ((a) \& (d)). However, we see interesting differences at finer levels (implications of protagonist's actions on others and vice versa). For example, female protagonist are more likely to make others feel positive (Figure \ref{fig:VA}(c)), and male protagonists are more aroused as a result of others' behaviors (Figure ~\ref{fig:VA}(e)). 



\begin{figure*}[t!]
\centering
\includegraphics[width=.85\linewidth]{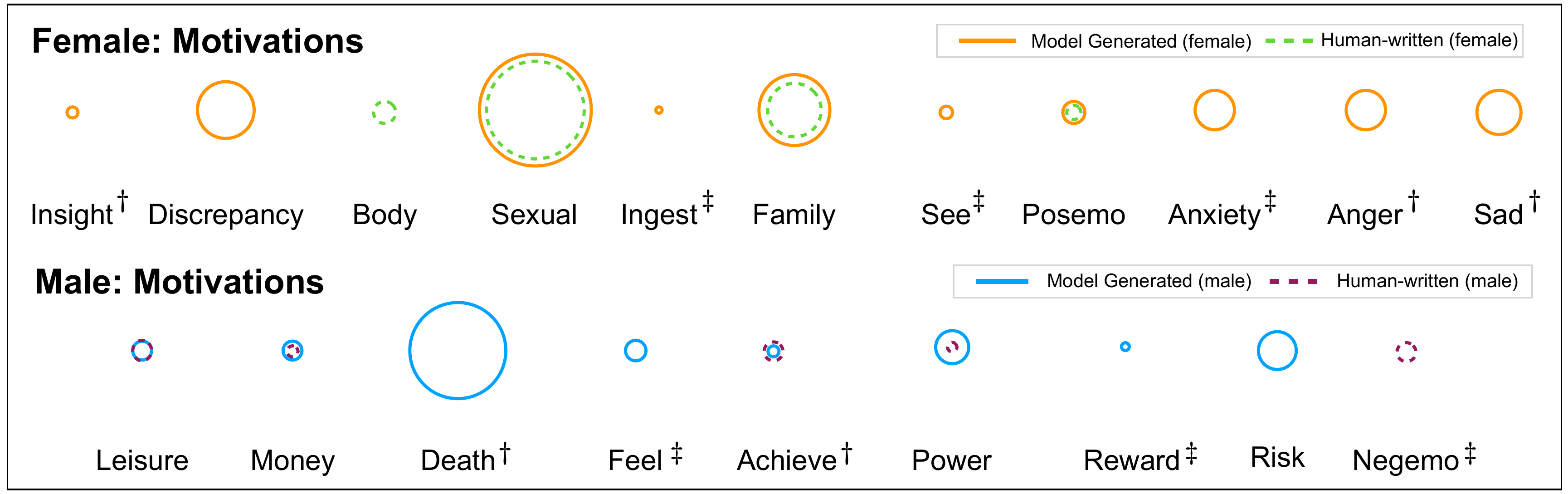}
\vspace{-5pt}
\caption{LIWC categories correlated with gender. The bigger the circles, the higher the correlations. All results are statistically significant at $p < 0.001$, except $^\dag p < 0.01$, and $^\ddag p < 0.05$.}
\label{fig:motiv}
\end{figure*}

\subsection{Motivations}

We now explore whether male and female protagonists have different motivations behind their actions. 
Having all motivation inferences (\pmot), we 
follow previous work by~\citet{rashkin-etal-2018-event2mind} and categorize \pmot into LIWC categories~\cite{Tausczik_thepsychological} based on their scores. 
For our analysis, we only consider `Core Drives and Needs', `Biological Process', `Personal Concerns', 'Perceptual Process', and `Social and Affect Words'.\footnote{For list of category descriptions please refer to: \url{http://liwc.wpengine.com/compare-dictionaries/}} We conduct regression analysis using Generalized Linear Model to obtain the correlations between gender and each LIWC category.\footnote{We statistically control for total number of words to account for gender skew in stories.}

As shown in Figure~\ref{fig:motiv}, female protagonists tend to have discrepancy, body, sexual, and family-related motivations, whereas male protagonists' actions are motivated by leisure, money, power, risk, and violence (death). Note that some categories (e.g. Anxiety, Death, and Risk) do not show correlations with gender in human-written stories, but do so in automatically generated stories.

\subsection{Classification based on bias}
To further quantify the significance of the implicit gender bias in GPT-2 generated stories, we train a classifier to predict the gender on story-level given all social inferences about \pat, \pem, \pemo, and \pmot 
(see \S\ref{subsec::social_imp}).
We fine-tune a pre-trained BERT-base model~\cite{devlin-etal-2019-bert} as our gender classification model. 
The classifier takes all implications (concatenated by \texttt{[SEP]} token) as input and the gender of the story's protagonist as output.
This classifier achieves an accuracy of $68.15\%$ on the test set. This indicates the implicit gender bias in GPT-2 generated stories is significant enough to leak the gender information.

\section{Conclusion}

Automatic story generation has real-world applications in entertainment, training and educating children. Hence, it is important for the generated stories to be socially unbiased. While biases can be expressed both explicitly and implicitly, this paper highlights the \textit{implicit} gender bias in automatically generated stories. We devised a pipeline to uncover implicit gender bias about the protagonist in model-generated stories using a commonsense inference engine. We show that male and female protagonists are portrayed with certain stereotypes: male protagonists are portrayed as more intellectual, while female protagonists are portrayed as more sexual and beautiful. In terms of mental states, female protagonists are more positive than male protagonists when interacting with others. 
Finally, we found protagonists' motives to be gendered as well. 

Our method can be used during post-hoc analysis of automatic story generation systems to quantify the genderness of their generated stories. Also, while designing gender-neutral models is out of the scope of the current paper, future work can use our findings to design unbiased story generators such as by using our experiments to design rewards for Reinforcement Learning models.

Lastly, following prior work, we analyze gender bias in a binary gender setup. A more realistic analysis, left for future work, should consider gender as a spectrum. 
We hope our study will encourage future work to devise methods for mitigating not just explicit but also \textit{implicit} gender biases.

\section*{Acknowledgements}
We would like to thank the anonymous reviewers for their helpful comments.

\section{Ethics Statement}

To generate stories for our analysis, we use a language model, GPT-2, pre-trained on WebText which has been shown to have potential harms and misuses~\cite{BenderGMS21}. The inductive bias of our fine-tuned model can limit the negative impacts to some extent. However, such pitfall motivates us to specifically examine the implicit gender bias in generated stories in more depth. We performed our experiments on a binary-gender setup to scale the scope of our analysis, while we acknowledge that these two groups are unrepresentative of real-world diversity. We hope our work and findings motivate future work to carefully design debiasing systems, making NLP systems safer and more equitable.

\bibliography{references}
\bibliographystyle{acl_natbib}




\end{document}